\documentclass{article}
\usepackage{spconf,amsmath,graphicx,mathrsfs}
\usepackage[colorlinks, citecolor=green]{hyperref}
\usepackage{amsfonts}
\usepackage{epstopdf}

\title{ Topic-Guided Attention for Image Captioning}
\name{Zhihao Zhu, Zhan Xue, Zejian Yuan}
\address{Institute of Artificial Intelligence and Robotics \\ School of Electronic and Information Engineering, Xi’an Jiaotong University
, P. R. China\\ \emph{\{704242527zzh@gmail.com\}, \{xx674967@stu.xjtu.edu.cn\}, \{yuan.ze.jian@xjtu.edu.cn\}}}
\begin{document}
%
\maketitle
\begin{abstract}
  Attention mechanisms have attracted considerable interest in image captioning because of its powerful performance. Existing attention-based models use feedback information from the caption generator as guidance to determine which of the image features should be attended to. A common defect of these attention generation methods is that they lack a higher-level guiding information from the image itself, which sets a limit on selecting the most informative image features. Therefore, in this paper, we propose a novel attention mechanism, called topic-guided attention, which integrates image topics in the attention model as a guiding information to help select the most important image features. Moreover, we extract image features and image topics with separate networks, which can be fine-tuned jointly in an end-to-end manner during training. The experimental results on the benchmark Microsoft COCO dataset show that our method yields state-of-art performance on various quantitative metrics.
\end{abstract}
\section{Introduction}
\label{sec:intro}
Automatic image captioning presents a particular challenge in computer vision because it needs to interpret from visual information to natural languages, which are two completely different information forms. Furthermore, it requires a level of image understanding that goes beyond image classification and object recognition. A widely adopted method for tackling this problem is the Encoder-Decoder framework\cite{NIC, C2, C3, C4}, where an encoder is first used to encode the pixel information into a more compact form, and later a decoder is used to translate this information into natural languages.

Inspired by the successful application of attention mechanism in machine language translation\cite{C7}, spatial attention has also been widely adopted in the task of image captioning. It's a feedback process that selectively maps a representation of partial regions or objects in the scene. On that basis, to further refine the spatial attention, some works\cite{C8, C9} applied stacked spatial attention, where the latter attention is based on the previous attentive feature map. Besides of spatial attention, Quanzeng et al\cite{C5} proposed to utilize the
\begin{figure}[t]
\begin{minipage}[b]{1.0\linewidth}
  \centering
  \centerline{\includegraphics[width=8.5cm]{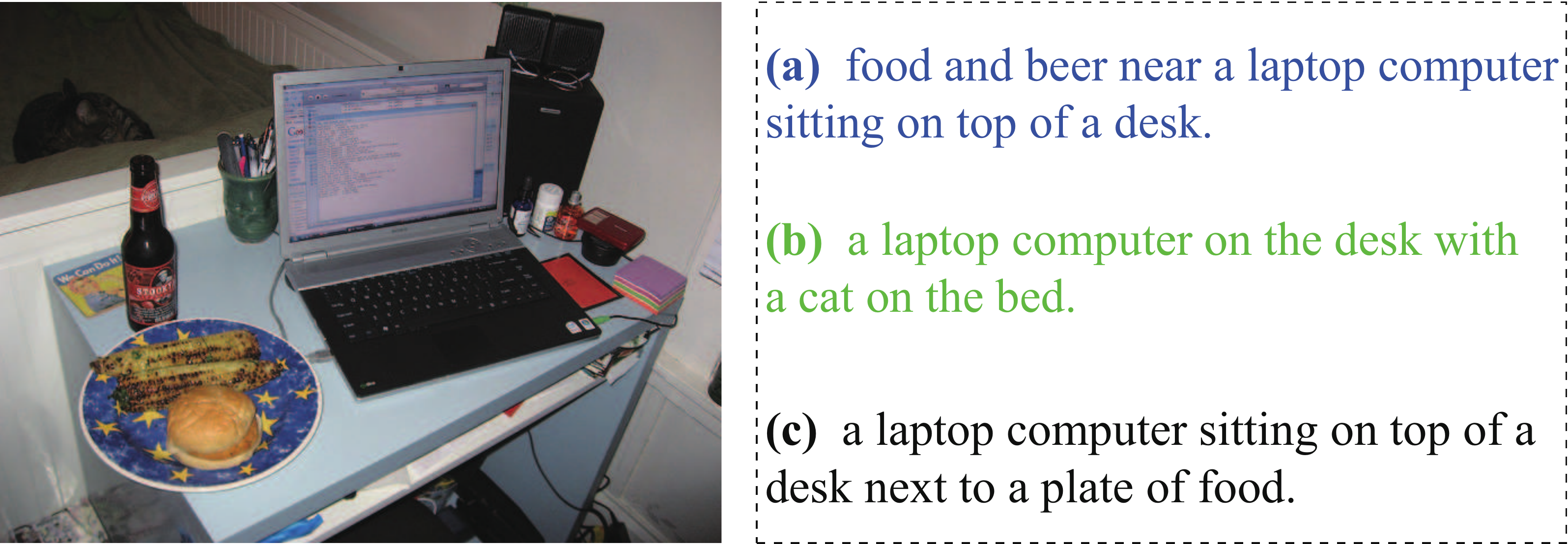}}
\end{minipage}

\caption{A comparison of captions generated from different methods. (a) represents for our proposed method, (b) stands for a baseline method\cite{C16} which does not use topic information. and (c) is the ground truth. }
\label{fig1}

\end{figure}
high-level semantic attributes and apply semantic attention to select the most important attributes at each time step.

However, a common defect of the above spatial and semantic attention models is that they lack a higher-level guiding information, which may cause the model to attend to some image regions that are visually salient but semantically irrelevant with image's main topic. In general, when describing an image, having an intuition about image's high-level semantic topic is beneficial for selecting the most semantically-meaningful and topic-relevant image areas and attributes as context for later caption generation.
For example, in Fig. \ref{fig1}, the image on the left depicts a scene where a laptop computer lies next to food. For we human, it's reasonable to infer the topic of the image to be ``working and eating". However, without this high-level guiding information, the baseline method \cite{C16} tends to describe all the salient visual objects in the image, including objects that are irrelevant to the general content of the image, such as the cat on the top left corner.

To solve the above issue, we propose a topic-guided attention mechanism that uses the image topic as a high-level guiding information. Our model starts with the extraction of image topic, based on image's visual appearance. Then, the topic vector is fed into the attention model together with the feedback from LSTM to generate attention on image visual features and attributes. The experimental results demonstrate that our method is able to generate captions that are more accordant with image's high-level semantic content.

The main contributions of our work consists of following two parts. \textbf{1)} we propose a new attention mechanism which uses image topic as auxiliary guidance for attention generation. The image topic acts like a regulator, maintaining the attention consistent with the general image content. \textbf{2)} we propose a new approach to integrate the selected visual features and attributes into caption generator. Our algorithm is able to achieve state-of-the-art performance on the Microsoft COCO dataset.

\section{Topic-guided attention for image captioning}
\label{sec:majhead}

\subsection{Overall Framework}
\begin{figure}[htb]
\begin{minipage}[b]{1.0\linewidth}
  \centering
  \centerline{\includegraphics[width=8.5cm]{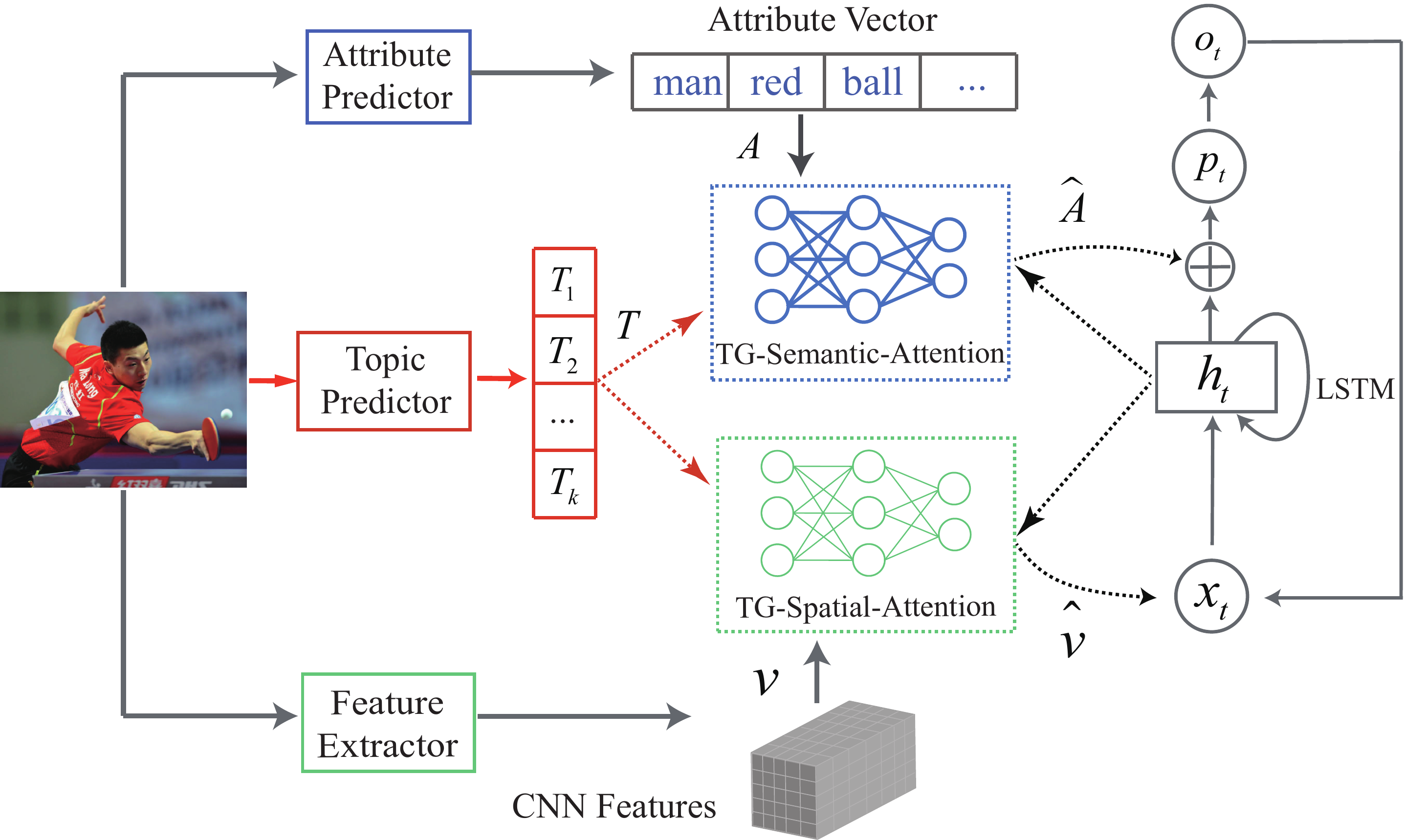}}
\end{minipage}

\caption{The framework of the proposed image captioning system. TG-Semantic-Attention represents the Topic-guided semantic attention model, and TG-Spatial-Attention represents the Topic-guided spatial attention model. }
\label{fig2}
\end{figure}

 Our topic-guided attention network for image captioning follows the Encoder-Decoder framework, where an encoder is first used for obtaining image features, and then a decoder is used for interpreting the encoded image features into captions. The overall structure of our model is illustrated in Fig. \ref{fig2}.

Different from other systems' encoding process, three types of image features are extracted in our model: visual features, attributes and topic. We use a deep CNN model to extract image's visual features, and a multi-label classifier, a single-label classifier are separately adopted for extracting image's topic and attributes (\emph{in section 3)}. Given these three information, we then apply topic-guided spatial and semantic attention to select the most important visual features  $\boldsymbol{\widehat v}$ and attributes $\boldsymbol{\widehat A}$ (\emph{Details in section 2.2 and 2.3)} to feed into decoder.

In the decoding part, we adopt LSTM as our caption generator. Different from other works, we employ a unique way to utilize the image topic, attended visual features and attributes: the image topic $T$ is fed into LSTM only at the first time step, which offers LSTM a quick overview of the general image content. Then, the attended visual features $\boldsymbol{\widehat v}$ and attributes $\boldsymbol{\widehat A}$ are fed into LSTM in the following steps. The overall working flow of LSTM network is governed by the following equations:

\begin{equation}\label{1}
{x_0} = {W^{x,T}}T
\end{equation}
\begin{equation}\label{2}
{x_t} = {W^{x,o}}{o_{t - 1}} \oplus {W^{x,v}}\boldsymbol{\widehat v}
\end{equation}
\begin{equation}\label{3}
{h_t} = LSTM({h_{t - 1}},{x_t})
\end{equation}
\begin{equation}\label{4}
{o_t} \sim {p_t} = \sigma (W^{o,h}{h_t}+W^{o,A}\widehat {\boldsymbol{A}})
\end{equation}
where $W$s denote weights, and $\oplus$ represents the concatenation manipulation. $\sigma$ stands for sigmoid function, $p_t$ stands for the probability distribution over each word in the vocabulary, and ${o_t}$ is the sampled word at each time step. For clearness, we do not explicitly represent the bias term in our paper.

\subsection{Topic-guided spatial attention}

In general, spatial attention is used for selecting the most information-carrying sub-regions of the visual features, guided by LSTM's feedback information. Unlike all previous works, we propose a new spatial attention mechanism, which integrates the image topic as auxiliary guidance when generating the attention.

We first reshape the visual features $\boldsymbol{\nu} = [ {\boldsymbol{v}_1},{\boldsymbol{v}_2},...,{\boldsymbol{v}_m}]$ by flattening its width $W$ and height $H$, where $m = W \cdot H$, and $\boldsymbol{v}_i\in{\mathbb R}^{D}$ corresponds to the \emph{i}-th location in the visual feature map. Given the topic vector $T$ and LSTM's hidden state $h_{t-1}$, we use a multi-layer perceptron with a softmax output to generate the attention distribution $\boldsymbol{\alpha} = \{\alpha_1, \alpha_2,...,\alpha_m\}$ over the image regions. Mathematically, our topic-guided spatial attention model can be represented as:
\begin{equation}\label{7}
  \boldsymbol{e} = {f_{MLP}}((W^{e,T}T) \oplus ({W^{e,\boldsymbol{\nu}}}\boldsymbol{\nu}) \oplus ({W^{e,h}}{h_{t - 1}}))
\end{equation}
\begin{equation}\label{8}
  \boldsymbol{\alpha}  = softmax({W^{\alpha,\boldsymbol{e}}}\boldsymbol{e})
\end{equation}
Where $f_{MLP}(\cdot)$ represents a multi-layer perceptron.

Then, we follow the ``soft'' approach to gather all the visual features to obtain $\widehat {\boldsymbol{v}}$ by using the weighted sum:
\begin{equation}\label{9}
\widehat {\boldsymbol{v}} = \sum\limits_{i = 1}^m {{\alpha _i}{\boldsymbol{v}_i}}
\end{equation}

\subsection{Topic-guided semantic attention}

Adding image attributes in the image captioning system was able to boost the performance of image captioning by explicitly representing the high-level semantic information\cite{C10}. Similar to the topic-guided spatial attention, we also apply a topic-guided attention mechanism on the image attributes $\boldsymbol{A} = \{A_1, A_2,..., A_n\}$, where $n$ is the size of our attribute vocabulary.

In our topic-guided semantic attention network, we use only one fully connected layer with a softmax to predict the attention distribution $\boldsymbol{\beta} = \{\beta_1, \beta_2,..., \beta_n\}$ over each attribute. The flow of the semantic attention can be represented as:
\begin{equation}\label{7}
  \boldsymbol{b} = {f_{FCL}}((W^{b,T}T) \oplus ({W^{b,A}\boldsymbol{A}}) \oplus ({W^{b,h}}{h_{t - 1}}))
\end{equation}
\begin{equation}\label{8}
 \boldsymbol{\beta}  = softmax({W^{\beta, \boldsymbol{b}}}\boldsymbol{b})
\end{equation}
where $f_{FCL}(\cdot)$ represents a fully-connected layer.

Then, we are able to reconstruct our attribute vector to obtain $\widehat {A}$ by multiplying each element with its weight:
\begin{equation}\label{11}
  \widehat {A}_i = {\beta _i} \odot {A_i},~~~~~~~~ \forall i \in n
\end{equation}
where $\odot$ denotes the element-wise multiplication, and $\widehat {A}_i$ is the $i$-th attribute in the $\widehat {A}$.

\subsection{Training}

Our training objective is to learn the model parameters by minimizing the following cost function:

\begin{equation}\label{12}
L =  - \frac{{\rm{1}}}{{\rm{N}}}\sum\limits_{i = 1}^N {\sum\limits_{t = 1}^{{L^{(i) + 1}}} {\log {p_t}(w_t^{(i)})} }  + \lambda  \cdot \left\| \theta  \right\|_2^2
\end{equation}
where $N$ is the number of training examples and $L^{(i)}$ is the length of the sentence for the $i$-th training example. ${p_t}(w_t^{(i)})$ corresponds to the Softmax activation of the $t$-th output of the LSTM, and $\theta$ represents model parameters, $\lambda  \cdot \left\| \theta  \right\|_2^2$ is a regularization term.

\section{Image Topic and Attribute Prediction}

\begin{table*}
\centering
\begin{tabular}{c|ccccccc}
\hline
 & BLEU-1 & BLEU-2 & BLEU-3 & BLEU-4 & METEOR & ROUGH-L & CIDEr-D \\ 
\hline
\hline
Google NIC\cite{NIC} & 66.6 & 46.1 & 32.9 & 24.6 & - & - & - \\
Soft attention\cite{C16} & 70.7 & 49.2 & 34.4 & 24.3 & 23.90 & - & - \\
Semantic attention\cite{C5} & 73.1 & 56.5 & 42.4 & 31.6 & 25.00 & 53.5 & 94.3 \\
PG-SPIDEr-TAG\cite{C20} & 75.1 & 59.1 & \textbf{44.6} & \textbf{33.6} & 25.5 & 55.1 & 104.2 \\
\hline
\hline
Ours-BASE & 74.8 & 55.8 & 41.1 & 30.2 & 27.0 & 57.8 & 109.8 \\
Ours-T-V & 75.2 & 56.16 & 41.4 & 30.4 & 27.0 & 58.1 & 109.2 \\
Ours-T-A & 75.8 & 57.0 & 42.5 & 30.9 & 27.4 & 58.2 & 112.5 \\
Ours-T-(V+A) & \textbf{77.8} & \textbf{59.34} & 44.5 & 33.2 & \textbf{28.60} & \textbf{60.1} & \textbf{117.6} \\

\end{tabular}
\caption{Performance of the proposed topic-guided attention model on the MSCOCO dataset, comparing with other four baseline methods.}
\vspace{-0.5em}
\label{table1}
\end{table*}

\textbf{Topic:}~~We follow \cite{C17} to first establish a training dataset of image-topic pairs by applying Latent Dirichlet Allocation (LDA) \cite{C18} on the caption data. Then, each image with the inferred topic label \emph{T} composes an image-topic pair. Then, these data are used to train a single-label classifier in a supervised manner. In our paper, we use the VGGNet\cite{C13} as our classifier, which is pre-trained on the ImageNet, and then fine-tuned on our image-topic dataset.\\
\\
\textbf{Attributes:}~~Similar to \cite{C11, C10}, we establish our attributes vocabulary by selecting $c$ most common words in the captions. To reduce the information redundancy, we perform a manual filtering of plurality (e.g. ``woman" and ``women") and semantic overlapping (e.g. ``child" and ``kid"), by classifying those words into the same semantic attribute. Finally, we obtain a vocabulary of 196 attributes, which is more compact than \cite{C11}. Given this attribute vocabulary, we can associate each image with a set of attributes according to its captions.

We then wish to predict the attributes given a test image. This can be viewed as a multi-label classification problem. We follow \cite{C19} to use a Hypotheses-CNN-Pooling (HCP) network to learn attributes from local image patches. It produces the probability score for each attribute that an image may contain, and the top-ranked ones are selected to form the attribute vector $\boldsymbol{A}$ as the input of the caption generator.

\section{Experiments}
In this section, we will specify our experimental methodology and verify the effectiveness of our topic-guided image captioning framework.
\subsection{Setup}

\textbf{Data and Metrics:}~~We conduct the experiment on the popular benchmark: Microsoft COCO dataset. For fair comparison, we follow the commonly used split in the previous works: 82,783 images are used for training, 5,000 images for validation, and 5,000 images for testing. Some images have more than 5 corresponding captions, the excess of which will be discarded for consistency. We directly use the publicly available code \footnote{https://github.com/tylin/coco-caption} provided by Microsoft for result evaluation, which includes BLEU-1, BLEU-2, BLEU-3, BLEU-4, METEOR, CIDEr, and ROUGH-L.\\
\textbf{Implementation details:}~~For the encoding part: \textbf{1)} The image visual features $\textbf{v}$ are extracted from the last 512 dimensional convolutional layer of the VGGNet. \textbf{2)} The topic extractor uses the pre-trained VGGNet connected with one fully connected layer which has 80 unites. Its output is the probability that the image belongs to each topic. \textbf{3)} For the attribute extractor, after obtaining the 196-dimension output from the last fully-connected layer, we keep the top 10 attributes with the highest scores to form the attribute vector $\boldsymbol{A}$.

For the decoding part, our language generator is implemented based on a Long-Short Term Memory (LSTM) network \cite{C14}. The dimension of its input and hidden layer are both set to 1024, and the \emph{tanh} is used as the nonlinear activation function. We apply a word embedding with 300 dimensions on both LSTM's input and output word vectors.

In the training procedure, we use Adam\cite{C15} algorithm for model updating with a mini-batch size of 128. We set the language model's learning rate to 0.001 and the dropout rate to 0.5. The whole training process takes about eight hours on a single NVIDIA TITAN X GPU.

\subsection{Quantitative evaluation results}
Table. \ref{table1} compares our method to several other systems on the task of image captioning on MSCOCO dataset. Our baseline methods inludes NIC\cite{NIC}, an end-to-end deep neural network translating directly from image pixels to natural languages, spatial attention with soft-attention\cite{C16}, semantic attention with explicit high-level visual attributes \cite{C5}. For fair comparison, we report our results with 16-layer VGGNet since it is similar to the image encoders used in other methods\cite{NIC, C5, C16}.

We also consider several systematic variants of our method: ( 1 ) OUR-BASE adds spatial attention and semantic attention jointly in the NIC model. ( 2 ) OUR-T-V adds image topic only to the spatial attention model in OUR-BASE. ( 3 ) OUR-T-A adds image topic only to the semantic attention model in OUR-BASE. ( 4 ) OUR-T-(V+A) adds image topic to both spatial and semantic attention in OUR-BASE.

On the MSCOCO dataset, using the same greedy search strategy, adding image topics to either spatial attention or semantic attention outperforms the base method (OUR-BASE) on all metrics. Moreover, the benefits of using image topics as guiding information in the spatial attention and semantic attention are addictive, proven by further improvement in OUR-T-(V+A), which outperforms OUR-BASE across all metrics by a large margin, ranging from 1\% to 5\%.

\subsection{Qualitative evaluations}
\vspace{-0.5em}

\begin{figure}[htb]

\begin{minipage}[b]{1.0\linewidth}
  \centering
  \centerline{\includegraphics[width=8.5cm]{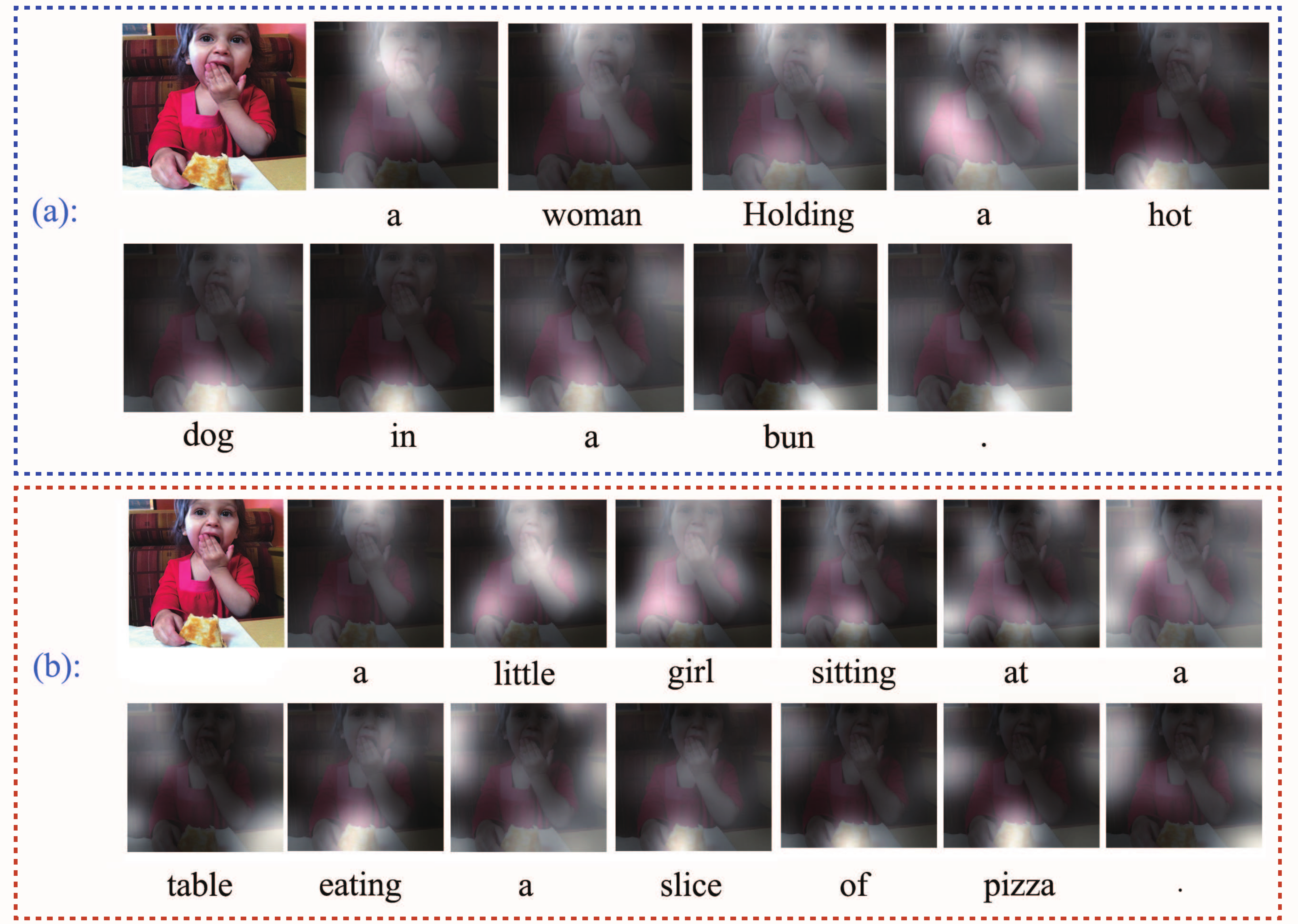}}

\end{minipage}

\caption{Example of generated spatial attention map and captions. \textbf{a)} OUR-BASE; ~ \textbf{b)} OUR-T-(V+A).}

\vspace{-1.5em}
\label{fig3}

\end{figure}
To evaluate our system qualitatively, in Fig. \ref{fig3}, we show an example demonstrating the effectiveness of topic-guided attention on the image captioning.

We note that first, the topic-guided attention shows a clearer distinction of object (the places where the attention would be focusing) and the background (where the attention weights are small). For example, when describing the little girl in the picture, our model gives a more precisely contoured attention areas covering the upper part of her body. In comparison, the base model pays the majority of attention to her head, while other body parts are overlooked.

Secondly, we observe that our model can better capture details in the target image, such as the adjective ``little" describing the girl and the quantifier ``a slice of" quantifying the ``pizza". Moreover, our model explores the spatial relation between the girl and the table: ``sitting at a table" which has even not been discovered by the baseline model. Also, the topic-guided attention discovers more accurate context information in the image, such as the verb ``eating" produced by our model, compared to the inaccurate verb ``holding" produced by the baseline model. This example demonstrates that topic-guided attention has a beneficial influence on the image caption generation task.

\section{Conclusion}
In this paper, we propose a novel method for image captioning. Different from other works, our method uses image topics as guiding information in the attention module to select semantically-stronger parts of visual features and attributes. The image topic in our model serves as two major functions: Macroscopically, it offers the language generator an overview of the high-level semantic content of the image; Microscopically, it's instructive for guiding the attention to exploit image's fine-grained local information.
For next steps, we plan to experiment with new methods to merge the spatial attention and semantic attention together in a single attention network.
\\
\\
\textbf{Acknowledgement: }
This work was supported by the National Key R\&D Program of China (No.2016YFB1001001) and the National Natural Science Foundation of China (No.91
\\
648121, No.61573280)

\bibliographystyle{IEEEbib}
\bibliography{refs}

\begin{thebibliography}{10}

\bibitem{NIC}
Oriol Vinyals, Alexander Toshev, Samy Bengio, and Dumitru Erhan,
\newblock ``Show and tell: Lessons learned from the 2015 {MSCOCO} image
  captioning challenge,''
\newblock {\em {IEEE} Trans. Pattern Anal. Mach. Intell.}, vol. 39, no. 4, pp.
  652--663, 2017.

\bibitem{C2}
Hanwang Zhang, Zawlin Kyaw, Shih{-}Fu Chang, and Tat{-}Seng Chua,
\newblock ``Visual translation embedding network for visual relation
  detection,''
\newblock in {\em 2017 {IEEE} Conference on Computer Vision and Pattern
  Recognition, {CVPR} 2017, Honolulu, HI, USA, July 21-26, 2017}, 2017, pp.
  3107--3115.

\bibitem{C3}
Andrej Karpathy and Li~Fei{-}Fei,
\newblock ``Deep visual-semantic alignments for generating image
  descriptions,''
\newblock {\em {IEEE} Trans. Pattern Anal. Mach. Intell.}, vol. 39, no. 4, pp.
  664--676, 2017.

\bibitem{C4}
Anna Rohrbach, Marcus Rohrbach, Ronghang Hu, Trevor Darrell, and Bernt Schiele,
\newblock ``Grounding of textual phrases in images by reconstruction,''
\newblock in {\em Computer Vision - {ECCV} 2016 - 14th European Conference,
  Amsterdam, The Netherlands, October 11-14, 2016, Proceedings, Part {I}},
  2016, pp. 817--834.

\bibitem{C7}
Dzmitry Bahdanau, Kyunghyun Cho, and Yoshua Bengio,
\newblock ``Neural machine translation by jointly learning to align and
  translate,''
\newblock {\em CoRR}, vol. abs/1409.0473, 2014.

\bibitem{C8}
Zichao Yang, Xiaodong He, Jianfeng Gao, Li~Deng, and Alexander~J. Smola,
\newblock ``Stacked attention networks for image question answering,''
\newblock in {\em 2016 {IEEE} Conference on Computer Vision and Pattern
  Recognition, {CVPR} 2016, Las Vegas, NV, USA, June 27-30, 2016}, 2016, pp.
  21--29.

\bibitem{C9}
Long Chen, Hanwang Zhang, Jun Xiao, Liqiang Nie, Jian Shao, Wei Liu, and
  Tat{-}Seng Chua,
\newblock ``{SCA-CNN:} spatial and channel-wise attention in convolutional
  networks for image captioning,''
\newblock in {\em 2017 {IEEE} Conference on Computer Vision and Pattern
  Recognition, {CVPR} 2017, Honolulu, HI, USA, July 21-26, 2017}, 2017, pp.
  6298--6306.

\bibitem{C5}
Quanzeng You, Hailin Jin, Zhaowen Wang, Chen Fang, and Jiebo Luo,
\newblock ``Image captioning with semantic attention,''
\newblock in {\em 2016 {IEEE} Conference on Computer Vision and Pattern
  Recognition, {CVPR} 2016, Las Vegas, NV, USA, June 27-30, 2016}, 2016, pp.
  4651--4659.

\bibitem{C16}
Kelvin Xu, Jimmy Ba, Ryan Kiros, Kyunghyun Cho, Aaron~C. Courville, Ruslan
  Salakhutdinov, Richard~S. Zemel, and Yoshua Bengio,
\newblock ``Show, attend and tell: Neural image caption generation with visual
  attention,''
\newblock in {\em Proceedings of the 32nd International Conference on Machine
  Learning, {ICML} 2015, Lille, France, 6-11 July 2015}, 2015, pp. 2048--2057.

\bibitem{C10}
Qi~Wu, Chunhua Shen, Lingqiao Liu, Anthony~R. Dick, and Anton van~den Hengel,
\newblock ``What value do explicit high level concepts have in vision to
  language problems?,''
\newblock in {\em 2016 {IEEE} Conference on Computer Vision and Pattern
  Recognition, {CVPR} 2016, Las Vegas, NV, USA, June 27-30, 2016}, 2016, pp.
  203--212.

\bibitem{C20}
Siqi Liu, Zhenhai Zhu, Ning Ye, Sergio Guadarrama, and Kevin Murphy,
\newblock ``Optimization of image description metrics using policy gradient
  methods,''
\newblock {\em CoRR}, vol. abs/1612.00370, 2016.

\bibitem{C17}
Kun Fu, Junqi Jin, Runpeng Cui, Fei Sha, and Changshui Zhang,
\newblock ``Aligning where to see and what to tell: Image captioning with
  region-based attention and scene-specific contexts,''
\newblock {\em {IEEE} Trans. Pattern Anal. Mach. Intell.}, vol. 39, no. 12, pp.
  2321--2334, 2017.

\bibitem{C18}
Thomas~G. Dietterich, Suzanna Becker, and Zoubin Ghahramani, Eds.,
\newblock {\em Advances in Neural Information Processing Systems 14 [Neural
  Information Processing Systems: Natural and Synthetic, {NIPS} 2001, December
  3-8, 2001, Vancouver, British Columbia, Canada]}. {MIT} Press, 2001.

\bibitem{C13}
Karen Simonyan and Andrew Zisserman,
\newblock ``Very deep convolutional networks for large-scale image
  recognition,''
\newblock {\em CoRR}, vol. abs/1409.1556, 2014.

\bibitem{C11}
Hao Fang, Saurabh Gupta, Forrest~N. Iandola, Rupesh~Kumar Srivastava, Li~Deng,
  Piotr Doll{\'{a}}r, Jianfeng Gao, Xiaodong He, Margaret Mitchell, John~C.
  Platt, C.~Lawrence Zitnick, and Geoffrey Zweig,
\newblock ``From captions to visual concepts and back,''
\newblock in {\em {IEEE} Conference on Computer Vision and Pattern Recognition,
  {CVPR} 2015, Boston, MA, USA, June 7-12, 2015}, 2015, pp. 1473--1482.

\bibitem{C19}
Yunchao Wei, Wei Xia, Junshi Huang, Bingbing Ni, Jian Dong, Yao Zhao, and
  Shuicheng Yan,
\newblock ``{CNN:} single-label to multi-label,''
\newblock {\em CoRR}, vol. abs/1406.5726, 2014.

\bibitem{C14}
Sepp Hochreiter and J{\"{u}}rgen Schmidhuber,
\newblock ``Long short-term memory,''
\newblock {\em Neural Computation}, vol. 9, no. 8, pp. 1735--1780, 1997.

\bibitem{C15}
Diederik~P. Kingma and Jimmy Ba,
\newblock ``Adam: {A} method for stochastic optimization,''
\newblock {\em CoRR}, vol. abs/1412.6980, 2014.

\end{thebibliography}
\end{document}